# SpaceYOLO: A Human-Inspired Model for Real-time, On-board Spacecraft Feature Detection

**Trupti Mahendrakar**
Florida Institute of Technology
150 W. University Blvd
Melbourne, FL 32901
tmahendrakar2020@my.fit.edu

**Ryan T. White**
Florida Institute of Technology
150 W. University Blvd.
Melbourne, FL 32901
rwhite@fit.edu

**Markus Wilde**
Florida Institute of Technology
150 W. University Blvd.
Melbourne, FL 32901
mwilde@fit.edu

**Madhur Tiwari**
Florida Institute of Technology
150 W. University Blvd.
Melbourne, FL 32901
mtiwari@fit.edu

*Abstract*—The rapid proliferation of non-cooperative spacecraft and space debris in orbit has precipitated a surging demand for on-orbit servicing and space debris removal at a scale that only autonomous missions can address, but the prerequisite autonomous navigation and flightpath planning to safely capture an unknown, non-cooperative, tumbling space object is an open problem. This requires algorithms for real-time, automated spacecraft feature recognition to pinpoint the locations of collision hazards (e.g. solar panels or antennas) and safe docking features (e.g. satellite bodies or thrusters) so safe, effective flightpaths can be planned. Prior work in this area reveals the performance of computer vision models are highly dependent on the training dataset and its coverage of scenarios visually similar to the real scenarios that occur in deployment. Hence, the algorithm may have degraded performance under certain lighting conditions even when the rendezvous maneuver conditions of the chaser to the target spacecraft are the same. This work delves into how humans perform these tasks through a survey of how aerospace engineering students experienced with spacecraft shapes and components recognize features of the three spacecraft: Landsat, Envisat, Anik, and the orbiter Mir. The survey reveals that the most common patterns in the human detection process were to consider the shape and texture of the features—antenna, solar panels, thrusters, and satellite bodies. This work introduces a novel algorithm SpaceYOLO, which fuses a state-of-the-art object detection algorithm YOLOv5 with a separate neural network based on these human-inspired decision processes exploiting shape and texture. Performance in autonomous spacecraft detection of SpaceYOLO is compared to ordinary YOLOv5 in hardware-in-the-loop experiments under different lighting and chaser maneuver conditions at the ORION Laboratory at Florida Tech.

## TABLE OF CONTENTS

## 1. INTRODUCTION

With the sharp rise in near-earth space debris, autonomous On-Orbit Servicing (OOS) and Active Debris Removal (ADR) operations around non-cooperative resident space object (RSO) has seen renewed interest.

Currently, the US Department of Defense's global Space Surveillance Network (SSN) tracks more than 27,000 space debris that are greater than 10cm and around a third of them are old, decommissioned spacecraft. An effective approach to prevent further growth of debris is by deorbiting or servicing decommissioned spacecraft. This prevents the risk of smaller debris colliding with them and forming more debris.

Some related OOS and ADR missions that demonstrated rendezvous, capture or inspection around a resident space object include ETS-VII [1], NASA, DARPA and AFRLs DART [2], XSS-10 [3], XSS-11 [4], ANGELS [5], MiTEx, Orbital Express [6], Northrop Grumman's MEV-1/-2 [7, 8], and Astroscale's ELSA-d [9].

These missions involved a single chaser spacecraft approaching a cooperative RSO with known geometry, capture interfaces, stable attitude, or ability to estimate stable attitude with RSOs surface markings.

Though the missions demonstrated that it is feasible to perform autonomous operations around a cooperative spacecraft, autonomous operations around a non-cooperative spacecraft, representative of space debris is still an unresolved challenge.

Our research at the ORION facility focuses on conducting autonomous ADR and OOS operations with a swarm of chasers around a non-cooperative spacecraft whose structure, functionality and attitude are unknown like the non-cooperative RSOs that are currently in orbit.

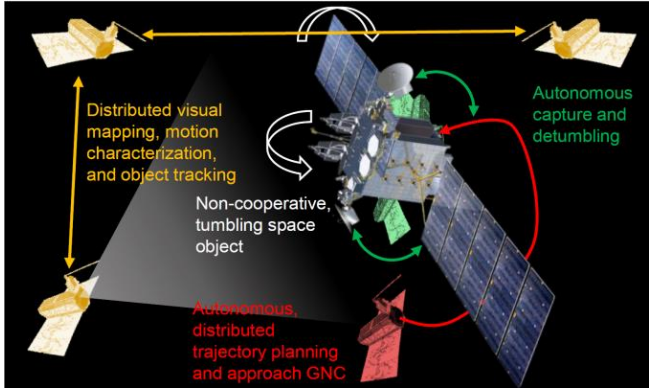

**Figure 1: Autonomous Docking with Swarm Satellites**

As described in [10, 11, 12], we use machine vision with YOLOv5 models on raspberry pi 4 to classify and locate components of non-cooperative spacecraft such as body, thrusters, solar panels, and antennas. Based on the position and attitude data of these components, an artificial potential field guidance algorithm evaluates a safe approach trajectory for our swarm of chaser spacecraft (DJI Tello Talent Drones) to autonomously navigate to the non-cooperative spacecraft.

Our previous machine vision algorithms – YOLOv5 and Faster RCNN were trained on over 1200 images of spacecraft from web searches and synthetic data. These images were augmented based on certain criteria to artificially increase the training data. Despite various training efforts, the machine vision algorithms still struggled to accurately detect components. To address this issue, we propose a new method called SpaceYOLO that uses contextual understanding of spacecraft to improve detection. This paper is written to serve as proof of concept of SpaceYOLO and the method requires further research and development prior to implementation.

The paper is structured as follows: sections 2 and 3 consist of literature review on YOLO and context-based learning along with the survey conducted to develop the initial version of SpaceYOLO. Sections 4 and 5 discuss the development of SpaceYOLO and the training dataset. Finally, the paper concludes with a discussion of the results of the proof of concept and future work.

## 2. COMPUTER VISION IN SPACE

Convolutional neural networks (CNNs) accelerated by graphics processing units (GPUs) [15] have revolutionized the performance of computer vision over the past 10 years. Initially, computer vision models required conventional computing hardware to supply sufficient computational power and huge visual datasets to train them. More recent innovations in small computing components, particularly edge GPUs (e.g. Intel Neural Compute Stick or NVIDIA Jetson) and FPGAs (e.g. Xilinx Kria KV260), have tremendously increased computational power available on low energy budgets making it feasible to be used on spacecraft [16, 17]. Additionally, there have been innovations in CNN architectures optimized for deployment on edge hardware [18, 19].

In the computer vision literature, the problem this paper is addressing falls into the area of object detection, which combines locating objects by drawing a bounding box tightly surrounding each object and classifying each as one of a pre-determined set of object types. From an input image, an object detector will ideally output three things for each object in the image: (1) a bounding box as the pixel position coordinates of the center of the box with width and height, (2) objectness score measuring how confident the model is that the box contains an object, and (3) a predicted object class.

There are three major types of object detection algorithms in the literature: vision transformer-based detectors, multi-stage detectors, and single-stage detectors. The computational cost of transformers rules them out for on-board in-space use. Multi-stage detectors attack the object detection problem in multiple stages, most frequently using one stage of the model to propose regions where objects *might* exist and another stage to classify those regions. Single-stage detectors do it all in one shot—they map image pixels directly to bounding box estimates, confidences, and classifications with a single neural network.

A multi-stage detector (Faster R-CNN) was previously shown to deliver higher accuracy than a single-stage detector (YOLOv5) at satellite component detection [14]. However, the framerate of Faster R-CNN is too low (approximately 0.2 FPS) on the proposed hardware, which would be unsafe for guiding a chaser to a tumbling satellite. Further, YOLOv5 produces good accuracy at a much higher framerate, reaching 2-5 FPS. Although the quality of the detections is slightly diminished, they occur 10-25 times faster, meaning an occasional missed detection is not as risky since it will be corrected very quickly in later frames.

A major disadvantage of typical object detectors is that they are not very interpretable, verging on black boxes. This article has taken a different approach to object detection that exploits the efficiency of YOLOv5, and the thought processes humans employ to perform the same task of identifying satellite components and drawing bounding boxes around them. Through surveys of the task performed by humans, we find common thought processes and build an object detector that uses these ideas explicitly to make decisions on classifications.

## 3. CONTEXT BASED DETECTION AND SURVEY OF HUMAN DETECTION

As previously mentioned, the idea behind using context-based detection techniques in SpaceYOLO is to minimize the detection issues found in our previous research in [10, 14, 12, 11]. A common issue we noticed was that the algorithm would misclassify under varying lighting conditions. Such misclassifications would be avoidable if the algorithm were more aware of the fundamentals of what these target components looked like and their fundamental use on the spacecraft, like how humans think.

Hence, the goal of SpaceYOLO is to force human-interpretable information based on visual context distantly related to the work conducted in [27]. In [27], the authors simplify an object recognition algorithm to recognize and categorize the place they are in and object around them. They identify the overall type of scene and then identify objects within the scene.

Some other works that inspire the SpaceYOLO model presented in this paper are from [28, 29, 30] where the authors correlate spatial information, scaling, textural information with the surroundings to classify the object.

*Survey*

To identify the criteria for the context-based technique, a survey was conducted among 24 aerospace engineering students, familiar with spacecraft components. The survey was used to better understand how humans use geometric shapes, colors and context to identify and classify features.

Each student was presented with the same four distinct spacecraft images – Landsat, Envisat, Anik, and Mir. They were asked to identify spacecraft components such as solar panels, body, antennas (horn, parabolic, phased array), thrusters and radiators. In addition to identifying those features, they were also required to provide their reasoning for each feature. The survey images with overall results are marked in Figure 2 through Figure 5 below. The top three reasoning for features - solar panels, parabolic antennas, thrusters, and body are tabulated in Table 1. These reasonings are used to form the basis of SpaceYOLO.

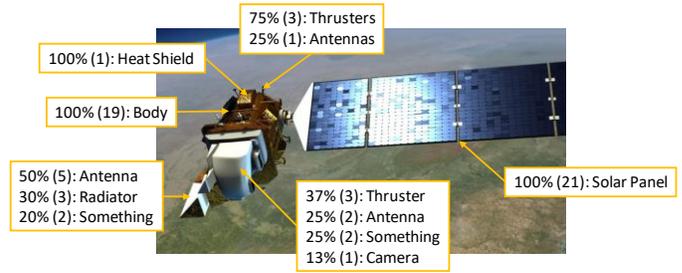

**Figure 2: Landsat Survey Results**

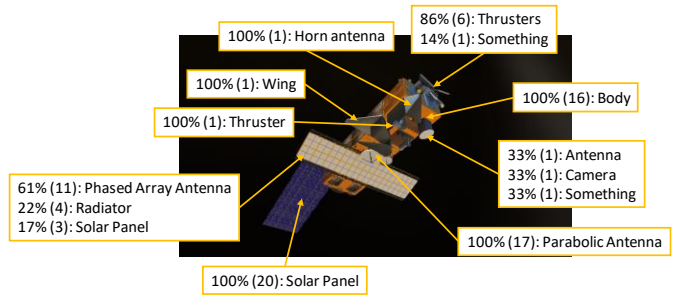

**Figure 3: Envisat Survey Results**

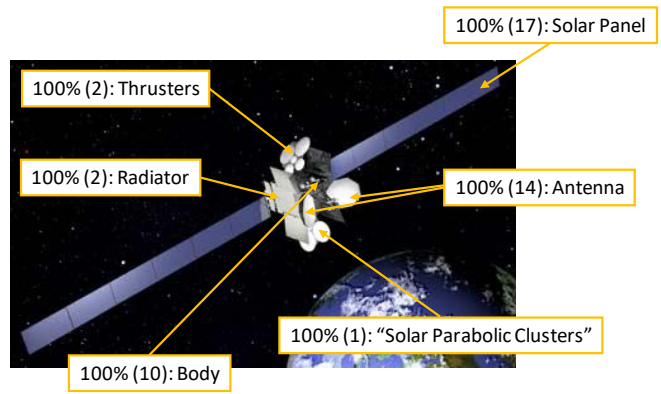

**Figure 4: Anik Survey Results**

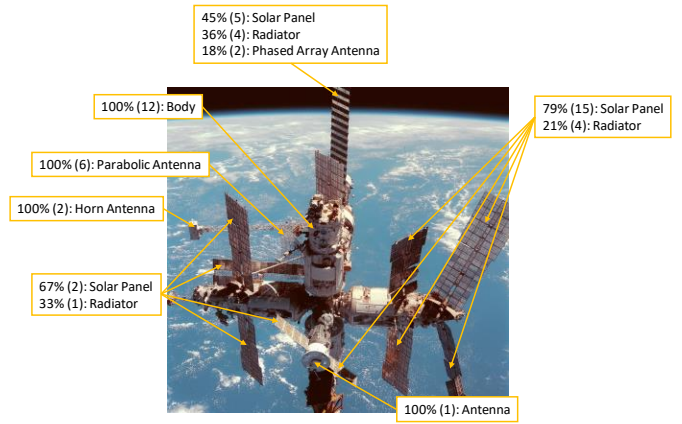

**Figure 5: Orbiter Mir Survey Results**

**Table 1: Common Reasoning Occurrences**

| Solar Panel | Antennas | Body | Thrusters |
|---|---|---|---|
| Rectangular | Circular, concave, Disk | Cuboid or Cylindrical | Conical |
| Reflective | Metallic | Covered in non-descriptive features (high visual density) | Metallic, shiny but not reflective |
| Thin Profile protruding from away from the body | Protruding from Body | Largest Single Object | Directly on body |

From Table 1, it is evident that the first step to identify the components is to detect shapes. The second step in the process is texture of the components, and the third is the spatial position of the features. Based on this interpretation, the SpaceYOLO version in this paper only employs steps 1 and 2 to study the concept's feasibility.

## 4. SPACEYOLO

Leveraging the data accrued from the survey, the first portion of the current version of SpaceYOLO consists of shape detector, and the second portion consists of variance classifier. The shape detector looks for circular and rectangular shapes as listed in Table 1. The variance classifier classifies these detected shapes from shape detector based on texture. The texture of the shapes is evaluated based on the pixel variance in the grayscale images.

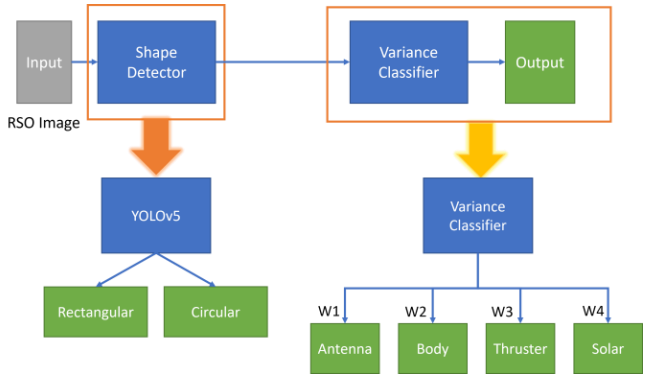

**Figure 6: SpaceYOLO Model**

*Shape Detector*

The shape detector in this algorithm is a YOLOv5s model trained to localize and classify shapes, specifically rectangles and circles. For the training dataset, a total of 30 images with circles, rectangles and triangles were built using MS Paint and labelled on Roboflow [31]. These images were augmented to artificially generate more images producing the final count of the training images to 53.

The augmentations include random rotation of images between ±45°, random shear between ±27° horizontally and vertically, random brightness adjustment between ±63% and, and salt and pepper noise for 5% of pixels. Below is a snippet of a few training images. These augmentations were picked based on the issues the camera might experience during tests and how relevant these shapes would be on a spacecraft.

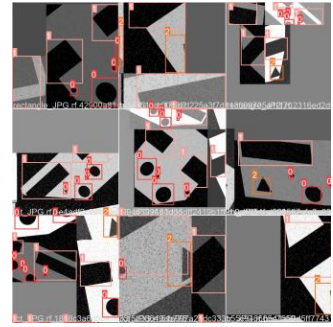

**Figure 7: Shape Detector Training Dataset**

Additionally, the training images were randomly placed with the goal of ensuring that the shapes covered most of the blank space within a training image frame. By doing this, the YOLOv5 algorithm will not be limited to a specific region within a test image to identify a shape. Below are the labeled heatmaps depicting that both classes properly covered most of the white space with these shapes.

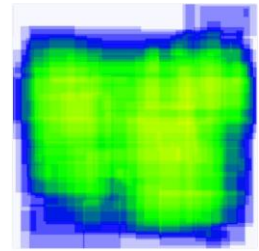

**Figure 8: Heatmap of All Classes**

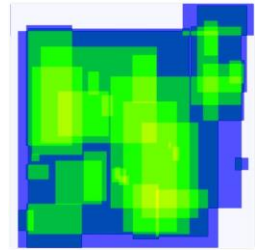 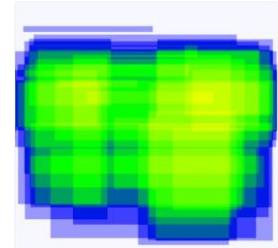

**Figure 9: Circle Heatmap**   **Figure 10: Rectangle Heatmap**

Since the idea behind the paper was not to define a benchmark for SpaceYOLO but introduce the concept of using human interpretability to identify the components, we were not concerned about finding the best weights for the model. Due to the limited dataset, training and comparing on the validation set was not an accurate representation of the detector. Therefore, we did not evaluate the weights on the validation dataset but rather tested it with hardware-in-the-loop experiments. Hence, for proof of concept, we started with something minimal that could detect simple shapes.

Additionally, we used neural network methods instead of edge detection because neural networks are capable of functioning with distortion, while edge detection cannot. For example, a neural network trained on rotated and sheared images of a rectangle would still be able to predict a rotated rectangular solar panel, which looks almost trapezoidal as a rectangle. However, edge detection would fail to identify this almost trapezoidal shape if it is tasked to look for rectangles.

*Variance Classifier*

The variance classifier is the second part of SpaceYOLO. The variance classifier was built based on the ORION Facility spacecraft mock-up model. Several videos of the spacecraft under extreme lighting conditions were taken while also changing the orientation

These videos were labelled at 1FPS, and each of the components (solar panel, body, thruster, and antenna images) were cropped out and converted to grayscale before extracting the pixel variance for each of the images. The histograms for each classes' variances for our lab data are shown in Figure 11. The variance value quantifies how much the pixels vary in the grayscale image.

Prior to using the spacecraft mock-up images from the lab, we extracted variances from synthetic image data. It consisted of images of spacecraft from the internet and NASA's stereolithography files. More information on this dataset is described in [12]. When using this varied set of images, it was noticed that the variance values were random leading to massive errors compared to the ones from hardware-in-the-loop images form the lab. This is a result of some images in the dataset coming from STK or the Kerbal Space Program, while others were from images of spacecraft in orbit or in a facility with non-realistic lighting conditions. Variance is directly related to lighting conditions and texture. Hence, for consistency and to obtain realistic variance values we chose lab images to build the variance classifier. For visualization, histograms of variances for each component for synthetic images are shown in Figure 12.

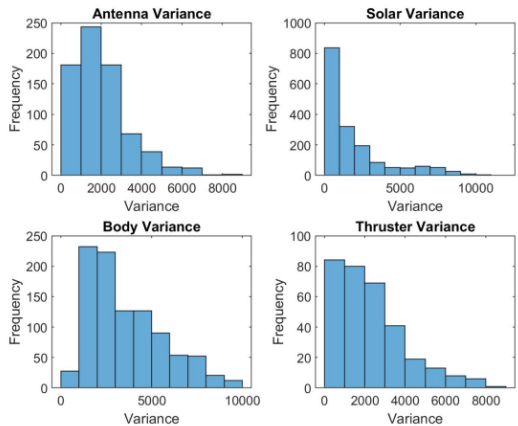

**Figure 11: Histogram of Variance – Lab Images**

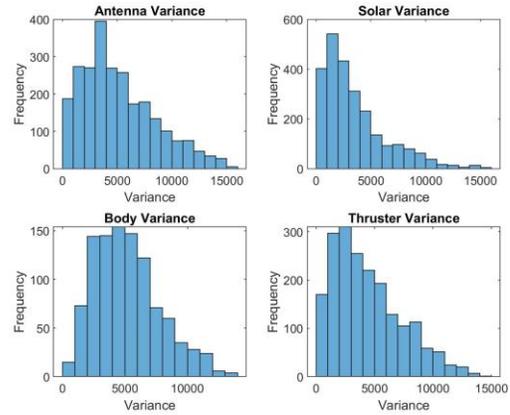

**Figure 12: Histogram of Variance - Synthetic Data**

There are two separate variance classifiers, one for each shape – rectangular and circular. Cropped images of the shapes from the shape detector pass into the respective variance classifier. The variance of the cropped image is compared to the histogram in Figure 11.

The variance classifier then outputs how probable the detected shape is an antenna, solar panel, thruster, or body and then gets multiplied by an optimized set of weighting factors depending on if it was detected as a circle or rectangle. The probability is calculated per Equation 1. $f_{c_n}$ is the number of images corresponding to class n in that variance range. The weights used in this paper for the circular branch are [0.45, 0.045, 0.45, 0.045] and rectangular branch are [0.2 0.2 0.2 0.4]. They were optimized by exhaustive search.

$$p(c_n) = \frac{f_{c_n}}{\sum_{i=1}^{4} f_{c_i}}, n = 1,2,3,4 \qquad (1)$$

## 5. TEST DATASET

As mentioned earlier, since the synthetic data was either too smooth or unrealistic, only hardware-in-the-loop experiment data was used for this paper. All the experimental data for the tests were collected from the ORION Facility at Florida Tech.

*ORION Facility*

The ORION facility [32] is equipped with a maneuver kinematics simulator, as seen in Figure 13. The simulator consists of a spacecraft mock-up. It is equipped with solar panels made from acrylic sheets, painted blue, the body is rectangular and wrapped with foil that accurately represents the Multilayer Insulation (MLI), the antenna is a smooth parabolic like dish, and finally, the thruster is conical shaped object, painted with low luster grey paint. Images of all the components are shown in Figure 14.

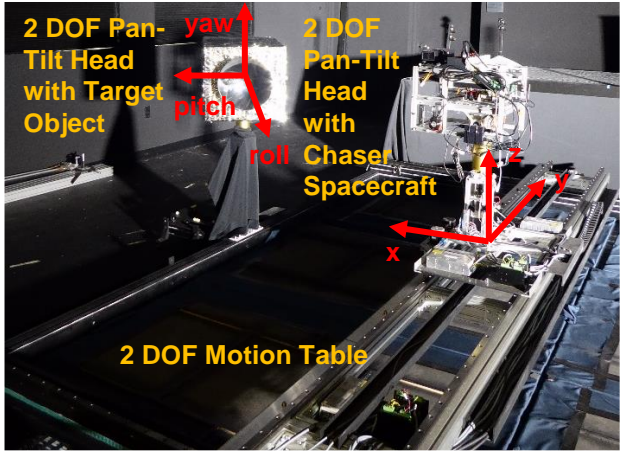

Figure 13: ORION Facility Florida Tech

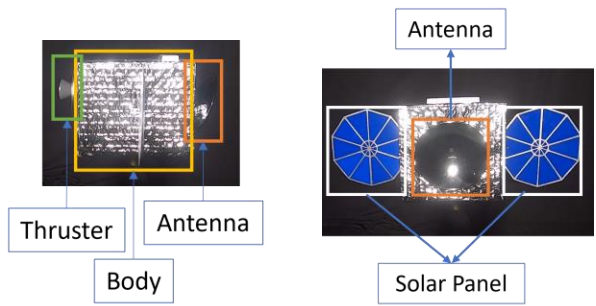

Figure 14: Spacecraft Components

The platform on which the mock-up is placed allows the spacecraft to rotate about three axes which creates yaw, pitch and roll at a maximum speed of $60°/s$ and maximum acceleration of $60°/s^2$.

The lighting is simulated with Hilio D12 LED Panel. The panel is rated for 350 W and for a color temperature of 5,600 K (daylight balanced). The intensity of the LED is equivalent to a 2,000 W incandescent lamp. The intensity can be continuously adjusted from 100% to 0% and vice versa. The beam angle can be varied between 10° and 60° using lens inserts, allowing us to replicate weaker and diffused Earth albedo effects and solar illumination.

Finally, the walls, ceiling and the floor of the facility are painted with low reflectivity black, and all the windows are covered with black-out blinds and black drapes to enable the simulation of realistic lighting conditions.

*Datasets*

A total of three datasets were studied to evaluate the performance of SpaceYOLO to compare it against YOLOv5 models developed in the past work [14] and against human predictions. The testing datasets shown in Figures 15, 16 and 17.

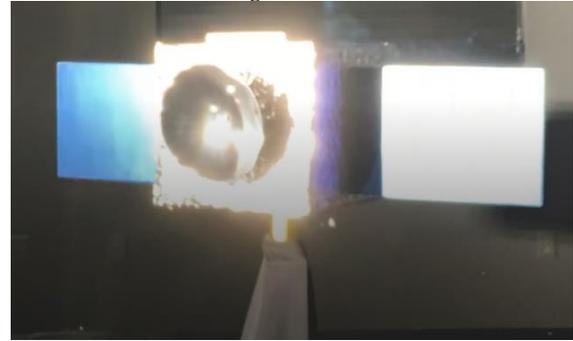

Figure 15: Test Dataset 1

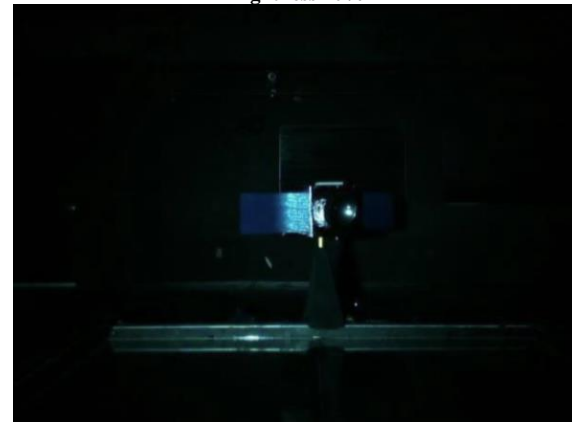

Figure 16: Test Dataset 2

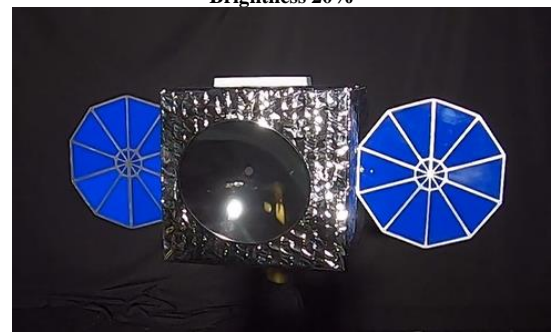

Figure 17: Test Dataset 3

### 6. EXPERIMENTS AND RESULTS

This section evaluates SpaceYOLO's shape detector and variance classifier's performance individually, and then compares it with YOLOv5s performance.

*SpaceYOLO Performance: Shape Detector*

SpaceYOLO's performance is collectively dependent on both the shape detector's and the variance classifier's performance.

To accurately depict the performance of the detector, we used precision and recall as our two-base metrics. Precision is described as the number of correct positive object predictions, while recall is the percentage of correctly classified ground truth objects at the class level. Both precision and recall are calculated using the following equations.

$$Precision\ (P) = \frac{TP}{TP + FP} \quad (2)$$

$$Recall\ (R) = \frac{TP}{TP + FN} \quad (3)$$

Following are the definitions of TP, FP and FN:
- True positives (TP): The correct classification of a ground-truth bounding box.
- False positives (FP): The incorrect classification of an object that is devoid or misplaced.
- False negative (FN): The missed classification of a ground-truth bounding box.

In order to have a well performing detector, it needs to have an increasing recall with high precision as the threshold of the confidence levels decreases. Average Precision (AP) depicts this relationship between precision and recall by utilizing a modified version of the area under a smoothed precision-recall curve for each class. Mean Average Precision(mAP) calculates the average AP value across all classes.

The shape detector's performance is evaluated based on mean average precision, mAP over an IoU (intersection of union) threshold of 0.5 and 0.5:0.95 as described in [12]. The shape detector's performance is evaluated based on mean average precision, mAP over an IoU (intersection of union) threshold of 0.5 and 0.5:0.95 as described in [12].

However, if the algorithm detects any component as a certain shape, the variance classifier further evaluates which class the shape belongs to. Hence, though we compare the mAP values, it does not control the outcome of SpaceYOLO.

The algorithm was fed in RGB images of hardware-in-the-loop datasets taken and labelled at 1FPS. For generalizing purposes, the antenna and thrusters were classified as circles while body and solar panels as rectangles. This forced any smooth edges to be detected as circles and sharp edges as rectangles.

The following Tables contain metrics on shape detector's performance for each dataset.

**Table 2: Test Dataset 1 Metrics**

| Class | Images | Instances | P | R | mAP@0.5 | mAP@.5:.95 |
|---|---|---|---|---|---|---|
| All | 30 | 100 | 0.481 | 0.37 | 0.31 | 0.0942 |
| Circle | 30 | 34 | 0.757 | 0.412 | 0.226 | 0.0548 |
| Rectangle | 30 | 66 | 0.684 | 0.327 | 0.394 | 0.133 |

*No detections in 4 out of 30 frames*

**Table 3: Test Dataset 2 Metrics**

| Class | Images | Instances | P | R | mAP@0.5 | mAP@.5:.95 |
|---|---|---|---|---|---|---|
| All | 81 | 308 | 0.278 | 0.268 | 0.192 | 0.0414 |
| Circle | 81 | 94 | 0.191 | 0.128 | 0.0493 | 0.0103 |
| Rectangle | 81 | 214 | 0.365 | 0.409 | 0.335 | 0.0724 |

*No detections in 34 out of 81 frames*

**Table 4: Test Dataset 3 Metrics**

| Class | Images | Instances | P | R | mAP@0.5 | mAP@.5:.95 |
|---|---|---|---|---|---|---|
| All | 42 | 159 | 0.46 | 0.53 | 0.5 | 0.20 |
| Circle | 42 | 52 | 0.27 | 0.67 | 0.54 | 0.17 |
| Rectangle | 42 | 107 | 0.64 | 0.38 | 0.45 | 0.23 |

*No detections in 11 out of 42 frames*

Dataset 1 performed the best out of the three classes due to the similarity of the training dataset with the spacecraft mockup. Additionally, the spacecraft was well illuminated, and the camera was closer to the mock-up compared to Dataset 2.

Dataset 2 performed the worst of all the three datasets since the images were taken from 5 meters which is the farthest out of the three cases. Additionally, to complicate the matter, the light was set to 20% which made it difficult to even label the video.

Notice that for Dataset 3 the solar panels are octagonal. Though the training dataset for the shape detector did not contain any octagons, the algorithm was still able to detect solar panels as rectangles while the sharp edges were prominent and detect as circles while the spacecraft was rotating, and the sharp edges were not prominent.

The outputs from the shape detector, cropped images of circles and rectangles shown in **Error! Reference source not found.** through Figure 21, were converted to grayscale and passed into the variance classifier.

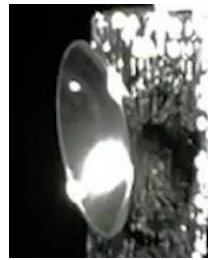

**Figure 18: Antenna detected as Circle**

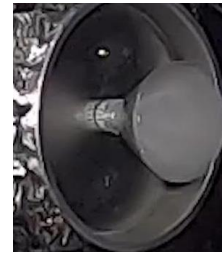

**Figure 19: Thruster Detected as Circle**

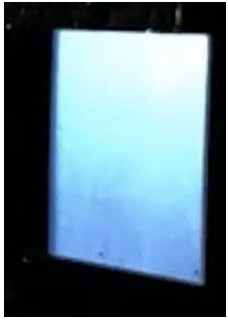 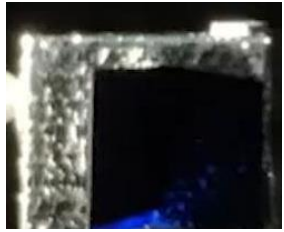

**Figure 20: Solar Panel detected as Rectangle**   **Figure 21: Solar Panel detected as Rectangle**

*SpaceYOLO Performance: Variance Classifier*

Currently, the performance of the variance classifier is evaluated manually. The gray scaled images from the shape detector were passed into the variance classifier where each image was either classified as body, solar panel, thruster, or antenna. Precision and recall for each of the two variance classifiers (rectangle and circle) for each dataset are tabulated as a confusion matrix and are discussed in this sub-section.

The rows of the confusion matrix correspond to the predicted values, the columns correspond to the ground truth or actual value. The number in the brackets under the component names in the top row corresponds to the actual number of component images either classified as circles or rectangles. The numbers in the diagonal of the matrix correspond to true positives, i.e., the number of images correctly classified as either antennas, body, thrusters, or solar panels. Other numbers along a specific row represent false positive (FP) i.e., for example 11 thrusters were falsely classified as antennas. Off-diagonal numbers along a column represent false negative (FN) values such as missed classifications. The right most column contains the precision for each class and the bottommost row contains the recall percentage for each class.

Per the APF algorithm as implemented in [13], thrusters and body are treated as attractive nodes i.e., the chaser spacecraft navigates to these points to dock with the RSO while avoiding solar panels and antennas. This means that higher recall values are preferred for thrusters and body while higher precision values are preferred for antennas and solar panels.

For dataset 1, the overall precision of the rectangular branch was 83.33% and the circular branch was 41.67%. It is seen that solar panels were classified with the highest precision and recall while thrusters performed with the lowest recall value.

**Table 5: Confusion Matrix for Test Dataset 1**

| | | Actual Value | | | | |
|---|---|---|---|---|---|---|
| | | Antenna (10) | Body (2) | Thruster (14) | Solar (10) | Precision |
| Predicted Value | Antenna | 8 | 0 | 11 | 0 | 42.11% |
| | Body | 0 | 1 | 1 | 1 | 33.33% |
| | Thruster | 1 | 0 | 2 | 0 | 66.67% |
| | Solar Panel | 1 | 1 | 0 | 9 | 81.82% |
| | Recall | 80% | 50% | 14.29% | 90% | |

For dataset 2, the precision of the rectangular branch was 72.72% and 51.28% for the circular branch. Overall, the shape detector for dataset 2 did not detect antennas and thrusters. This goes back to the reasoning that the pictures were taken from farther out requiring the shape detector to have been trained on highly zoomed out training image dataset. It was noticed that because the spacecraft was so far away and the height of both the body and solar panel were similar, both the solar panel and body combined were detected as a rectangle. When these cropped images were fed into the variance classifier, it was falsely detected as either solar panel or body. However, it is important to note that there were no detections for this dataset that were antenna or thruster indicating that the variance classifier behaves as expected.

**Table 6: Confusion Matrix Test Dataset 2**

| | | Actual Value | | | | |
|---|---|---|---|---|---|---|
| | | Antenna (0) | Body (26) | Thruster (0) | Solar (16) | Precision |
| Predicted Value | Antenna | 0 | 0 | 0 | 0 | N/A |
| | Body | 0 | 13 | 0 | 7 | 65% |
| | Thruster | 0 | 0 | 0 | 0 | N/A |
| | Solar Panel | 0 | 13 | 0 | 9 | 59% |
| | Recall | N/A | 50% | N/A | 56.25% | |

The accuracy of the variance classifier for dataset 3 for the rectangular branch was 57.143% and the circular branch was 42.87%. Though the solar panels were octagonal, the shape detector classified them as both circular and rectangular. Despite this bias, the variance classifier forced the final classification to be solar panels indicating that the variance classifier works.

**Table 7: Confusion Matrix for Test Dataset 3**

| | | Actual Value | | | | |
|---|---|---|---|---|---|---|
| | | Antenna (21) | Body (0) | Thruster (16) | Solar (13) | Precision |
| Predicted Value | Antenna | 19 | 0 | 15 | 4 | 50% |
| | Body | 0 | 0 | 1 | 0 | N/A |
| | Thruster | 0 | 0 | 0 | 0 | N/A |
| | Solar Panel | 2 | 0 | 0 | 9 | 81.82% |
| | Recall | 90.48% | N/A | 0% | 69.23% | |

Additionally, for a minimum of 7 instances, the variance classifier accurately forced the solar panel detections to 100% regardless of the shape detector results.

*YOLOv5 Experiments*

The performance of the YOLOv5 with weights from [12] are tested against the three datasets. The results are tabulated below. These results are explained in [14].

**Table 8: YOLOv5 Test Dataset 1**

| Class | Images | Instances | P | R | mAP@0.5 | mAP@.5:.95 |
|---|---|---|---|---|---|---|
| All | 30 | 98 | 0.35 | 0.37 | 0.32 | 0.27 |
| Antenna | 30 | 15 | 0.26 | 0.4 | 0.21 | 0.12 |
| Body | 30 | 30 | 0.60 | 0.5 | 0.52 | 0.2 |
| Solar | 30 | 36 | 0.31 | 0.28 | 0.32 | 0.18 |
| Thruster | 30 | 17 | 0.23 | 0.29 | 0.23 | 0.16 |

**Table 9: YOLOv5 Test Dataset 2**

| Class | Images | Instances | P | R | mAP@0.5 | mAP@.5:.95 |
|---|---|---|---|---|---|---|
| All | 81 | 308 | 0.93 | 0.31 | 0.38 | 0.18 |
| Antenna | 81 | 44 | N/A | 0 | 0.04 | 0.015 |
| Body | 81 | 81 | 0.89 | 0.60 | 0.71 | 0.34 |
| Solar | 81 | 133 | 0.82 | 0.60 | 0.60 | 0.27 |
| Thruster | 81 | 50 | 1 | 0.03 | 0.18 | 0.08 |

**Table 10: YOLOv5 Test Dataset 3**

| Class | Images | Instances | P | R | mAP@0.5 | mAP@.5:.95 |
|---|---|---|---|---|---|---|
| All | 42 | 159 | 0.385 | 0.364 | 0.328 | 0.242 |
| Antenna | 42 | 29 | 0.17 | 0.17 | 0.0945 | 0.0691 |
| Body | 42 | 42 | 0.639 | 0.639 | 0.353 | 0.181 |
| Solar | 42 | 65 | 0.462 | 0.462 | 0.635 | 0.574 |
| Thruster | 42 | 23 | 0.267 | 0.267 | 0.231 | 0.146 |

## 7. SPACEYOLO, YOLOV5 AND HUMAN DETECTION COMPARISON

Because this paper compares entirely different methods – SpaceYOLO, YOLOv5 and human detections, the testing metrics used to evaluate the performance is essentially the number of images localized and classified into different components out of all the images presented to the algorithm.

**Table 11: Results Comparison**

|  | Antenna | | Body | | Thruster | | Solar | |
|---|---|---|---|---|---|---|---|---|
|  | μ | β | μ | β | μ | β | μ | β |
| Set 1 | 8 | 7 | 1 | 14 | 2 | 3 | 9 | 18 |
| Labelled Occurrence | 15 | | 30 | | 17 | | 36 | |
| Set 2 | 0 | 0 | 13 | 53 | 0 | 5 | 9 | 176 |
| Labelled Occurrence | 44 | | 81 | | 50 | | 133 | |
| Set 3 | 19 | 24 | 0 | 8 | 0 | 12 | 9 | 57 |
| Labelled Occurrence | 29 | | 42 | | 23 | | 65 | |

μ: SpaceYOLO
β: YOLOv5

Results for each method are presented in Table 11. $\mu$ corresponds to SpaceYOLO, $\beta$ corresponds to YOLOv5 and Labelled Occurrence corresponds to the number of human labels, which forms the baseline. Based on results above, antennas in dataset 1 performed better than YOLOv5 however overall, YOLOv5 shows better performance. When compared to precision and recall values from the *SpaceYOLO: Variance Classifier* confusion matrices and YOLOv5 test metrics, SpaceYOLO shows promising results especially for identifying shapes of the spacecraft without ever being trained on spacecraft dataset and using only 53 training images instead of 1200+. If the shape detector was trained on more variety of shapes, the algorithm's overall accuracy would improve significantly since the results above show that the variance classifier has a huge impact on the final detections.

It is also seen that YOLOv5 surpassed human detections for recognizing solar panels for dataset 2. The algorithm was able to identify the slightest blue shade of solar panels despite the zoomed-out images of dataset 2. This shows the algorithm's ability to recognize colors and small shapes. The same could be implemented on SpaceYOLO by augmenting the image dataset to improve its ability to detect shapes. Additionally, since SpaceYOLO only detects shapes and evaluates variance to classify the detection (not a neural network), it significantly reduces the computational complexity otherwise required by the previously employed YOLOv5 methods.

The most current YOLOv5 model runs at 2FPS on Raspberry Pi [11] and, theoretically since SpaceYOLO requires less computational power, the algorithm should run at a higher frame rate, providing more reliable localization of spacecraft components.

## 8. CONCLUSIONS

The current work on SpaceYOLO demonstrates the benefits of bringing human decision-making processes into an AI framework. The benefits of this model include human interpretability which leads to less dependence on a black box to make safety-critical decisions. A secondary benefit of this type of implementation is generating and labelling shapes for SpaceYOLO can be easily automated. This tremendously reduces time and resource costs associated with data development.

Currently the algorithm demonstrates that this method is feasible but not quite up to par with the past YOLOv5 implementation however, SpaceYOLO has shown promising results that suggest future work in the direction of human interpretable computer vision implementation for OOS and ADR operations.

One of the major issues noticed during testing SpaceYOLO is that the images of antennas and thrusters contain part of the body. This leads to confusion in the variance classifier. Future work will evaluate the effects of disregarding the body and only detecting antennas, solar panels, and thrusters. In addition, the performance evaluation of the algorithm will be automated so testing with larger datasets can be feasible. Furthermore, the decision tree model for future work will expand more than variance and account for spatial position and orientation of the features.

## ACKNOWLEDGEMENTS

This project was supported by the NVIDIA Applied Research Accelerator Program. Additional conference funding support was provided by N000142012669 Inspiring Students to Pursue U.S. Navy STEM Careers through Experiential Learning grant. The authors wish to thank Dr. Brooke Wheeler for coordinating conference support. The authors also thank Mackenzie Meni whose help in editing the paper was critical to its submission.

# REFERENCES

[1] K. Yoshida, "Engineering Test Satellite VII Flight Experiments For Space Robot Dynamics and Control: Theories on Laboratory Test Beds Ten Years Ago, Now in Orbit," *The International Journal of Robotics Research,* vol. 22, no. 5, pp. 321-335, 2003.

[2] R. T. Howard and T. C. Bryan, "DART AVGS flight results," *Proceedings of SPIE,* vol. 6555, no. Sensors and Systems for Space Applications, pp. 1-10, 2007.

[3] T. M. Davis and D. Melanson, "XSS-10 Micro-Satellite Flight Demonstration Program Results," *Proc. of SPIE,* vol. 5419, no. Spacecraft Platforms and Infrastructure, pp. 16-25, 2004.

[4] AFRL, "XSS-11 Micro Satellite," September 2011. [Online]. Available: https://www.kirtland.af.mil/Portals/52/documents/AFD-111103-035.pdf?ver=2016-06-28-110256-797. [Accessed 17 July 2020].

[5] AFRL, "Automated Navigation and Guidance Experiment for Local Space (ANGELS)," July 2014. [Online]. Available: https://www.kirtland.af.mil/Portals/52/documents/AFD-131204-039.pdf?ver=2016-06-28-105617-297. [Accessed 17 July 2020].

[6] F. G. Kennedy III, "Orbital Express: Accomplishments and Lessons Learned," *Advances in the Astronautical Sciences,* vol. 131, no. Guidance and Control 2008, pp. 575-586, 2008.

[7] INTELSAT, "MEV-1: A look back at Intelsat's groundbreaking journey," 17 April 2020. [Online]. Available: https://www.intelsat.com/resources/blog/mev-1-a-look-back-at-intelsats-groundbreaking-journey/. [Accessed 2022 September 19].

[8] Space News, "MEV-2 servicer successfully docks to live Intelsat satellite," 12 April 2021. [Online]. Available: https://spacenews.com/mev-2-servicer-successfully-docks-to-live-intelsat-satellite/. [Accessed 19 September 2022].

[9] Astroscale, "Astroscale's ELSA-d Successfully Demonstrates Repeated Magnetic Capture," 25 August 2021. [Online]. Available: https://astroscale.com/astroscales-elsa-d-successfully-demonstrates-repeated-magnetic-capture/. [Accessed 19 September 2022].

[10] T. Mahendrakar, J. Cutler, N. Fischer, A. Rivkin, A. Ekblad, K. Watkins, M. Wilde, R. White and B. Kish, "Use of Artificial Intelligence for Feature Recognition and Flightpath Planning Around Non-Cooperative Resident Space Objects," in *ASCEND*, Las Vegas, 2021.

[11] T. Mahendrakar, S. Holmberg, A. Ekblad, E. Conti, R. T. White, M. Wilde and I. Silver, "Autonomous Rendezvous with Non-Cooperative Target Objects with Swarm Chasers and Observers," in *AAS Spaceflight Mechanics*, Austin, 2023.

[12] T. Mahendrakar, R. T. White, M. Wilde, B. Kish and I. Silver, "Real-time Satellite Component Recognition with YOLO-V5," in *Small Satellite Conference*, Utah, 2021.

[13] J. Cutler, M. Wilde, A. Rivkin, B. Kish and I. Silver, "Artificial Potential Field Guidance for Capture of Non-Cooperative Target Objects by Chaser Swarms," in *ASCEND 2021*, 2021.

[14] T. Mahendrakar, A. Ekblad, N. Fischer, R. White, M. Wilde, B. Kish and I. Silver, "Performance Study of YOLOv5 and Faster R-CNN for Autonomous Navigation around Non-Cooperative Targets," in *IEEE Aerospace Conference*, Montana, 2022.

[15] A. Krizhevsky, I. Sutskever and G. E. Hinton, "ImageNet Classification with Deep Convolutional Neural Networks," in *Advances in Neural Information Processing Systems*, 2012.

[16] N. Alarcon, "Technical Blog," Nvidia Developer, 07 August 2020. [Online]. Available: https://developer.nvidia.com/blog/lockheed-martin-usc-jetson-nanosatellite/. [Accessed 30 September 2022].

[17] XILINK, "RT Kintex UltraScale FPGAs for Ultra High Throughput and High Bandwidth Applications," WP523 (V1.0), May 19, 2020.

[18] A. G. Howard, M. Zhu, B. Chen, D. Kalenichenko, W. Wang, T. Weyand, M. Andreetto and H. Adam, "MobileNets: Efficient Convolutional Neural Networks for Mobile Vision Applications," in *arXiv preprint arXiv:1704.04861*, 2017.

[19] J. Redmon, S. Divvala, R. Girshick and A. Farhadi, "You Only Look Once: Unified, Real-time Object Detection," in *IEEE Conference on Computer Vision and Pattern Recognition (CVPR)*, 2016.

[20] S. Stefano, M. Piccinin, G. Zanotti, A. Brandonisio, I. Bloise, L. Feruglio, P. Lunghi and M. Varile, "Optical navigation for Lunar landing based on Convolutional Neural Network crater detector," *Aerospace Science and Technology,* vol. 123, no. 1270-9638, p. 107503, 2022.

[21] P. Lunghi, M. Lavagna and R. Armellin, "A semi-analytical guidance algorithm for autonomous landing," *Advances in Space Research,* vol. 55, no. 0273-1177, pp. 2719-2738, 2015.

[22] L. Felicetti and R. M. Emami, "Vision-Aided Attitude Control for Space Debris Detection," *JGCD,* vol. 41, no. 2, pp. 573-575, 2018.

[23] J.-F. Shi, S. Ulrich and S. Ruel, "Spacecraft Pose Estimation using Principal Component Analysis and a Monocular Camera," *IAC,* no. 10.2514/6.2017-1034, 2014.

[24] S. Sharma and S. D'Amico, "Neural Network-Based Pose Estimation for Noncooperative Spacecraft

Rendezvous," *IEEE Transactions on Aerospace and Electronic Systems,* vol. 56, no. 6, pp. 4638-4658, 2020.

[25] A. Kenall, M. Gromes and R. Cipolla, "PoseNet: A Convolutional Network for Real-Time 6-DOF Camera Relocalization," *2015 IEEE International Conference on Computer Vision (ICCV),* no. 2380-7504, pp. 2938-2946, 2015.

[26] J. Song, D. Rondao and N. Aouf, "Deep learning-based spacecraft relative navigation methods: A survey," *Acta Astronautica,* vol. 191, pp. 22-40, 2022.

[27] A. Torralba, K. P. Murphy, W. T. Freeman and M. A. Rubin, "Context-based Vision System for Place and Object Recognition," *Proceedings Ninth IEEE International Conference on Computer Vision,* vol. 1, no. doi: 10.1109/ICCV.2003.1238354., pp. 273-280, 2003.

[28] J. Shotton, J. Winn and C. Rother, "TextonBoost for Image Understanding: Multi-Class Object Recognition and Segmentation by Jointly Modeling Texture, Layout, and Context," *International Journal Computer Vision,* vol. 81, no. DOI 10.1007/s11263-007-0109-1, pp. 2-23, 2009.

[29] M. Zhang, C. Tseng and G. Kreiman, "Putting visual object recognition in context," *Putting visual object recignition in context,* no. DOI: 10.1109/CVPR42600.2020.01300, pp. 12982-12991, 2020.

[30] L. Kondmann, A. Toker, B. Scholkopf, L. Leal-Taixe and X. x. Zhu, "Spatial Context Awareness for Unsupervised Change Detection in Optical Satellite Images," *IEEE Trasnactions on Geoscience and Remote Sensing,* vol. 60, no. DOI: 10.1109/TGRS.2021.3130842, 2022.

[31] "Roboflow," [Online]. Available: https://roboflow.com/. [Accessed 8 October 2021].

[32] M. Wilde, B. Kaplinger, T. Go, H. Guiterrez and D. R. Kirk, "ORION: A Teaching and Research Platform for Simulation of Space Proximity Operations," in *AIAA Space*, Pasadena, 2015.

## BIOGRAPHY

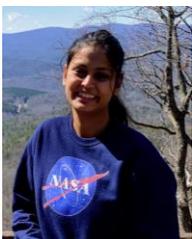

***Trupti Mahendrakar*** *received a B.S. in Aerospace Engineering from Embry-Riddle Aeronautical University, Prescott, Arizona in 2019 and a M.S, in Aerospace Engineering from Florida Institute of Technology, Melbourne. She is currently a Ph.D. candidate at the Florida Institute of Technology, Melbourne. Her current research includes implementation of machine vision algorithms to enhance on-orbit service satellite operations, optimization of cold gas thruster design, implementation of robotic manipulators for satellite refueling.*

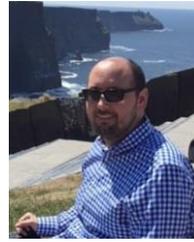

***Dr. Ryan T. White*** *is an Assistant Professor of Mathematics at Florida Tech and Director of the NEural TransmissionS (NETS) Lab. His research focuses on computer vision, physics-inspired machine learning, synthetic data generation, and probability. His work includes multidisciplinary projects on in-space use of computer vision to support on-orbit proximity operations, medical data analytics, glaciology, and geoinformation systems. Dr. White earned a Ph.D. from Florida Tech and joined the Florida Tech faculty in 2015.*

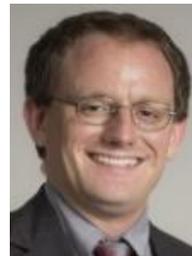

***Dr. Markus Wilde*** *is an Associate Professor for Aerospace Engineering at Florida Tech and Director of the ORION Lab. His research focus lies on experimental studies of spacecraft and aircraft control systems. Dr. Wilde received his M.S. and Ph.D. in Aerospace Engineering at TU Munich, Germany. He was accepted into the NRC Research Associateship Program in 2013, as postdoctoral associate at the Spacecraft Robotics Laboratory at the Naval Postgraduate School. In 2014, he joined the Florida Tech faculty.*

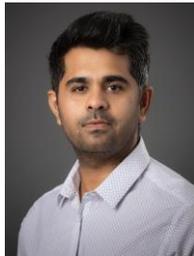

***Dr. Madhur Tiwari*** *is the Director of the Autonomy Lab and Assistant Professor of Aerospace Engineering at Florida Institute of Technology. He specializes in robotics, machine learning and control for aerospace systems. Currently, he is teaching Spaceflight Mechanics and Modern Control Theory at Florida Institute of Technology.*